\documentclass[letterpaper, 10 pt, conference]{ieeeconf}  

\IEEEoverridecommandlockouts                              
\usepackage[utf8]{inputenc}
\usepackage{amsmath,amssymb,stmaryrd,mathtools}
\usepackage{amsfonts}
\usepackage[noend]{algpseudocode}
\usepackage{acronym}
\usepackage{verbatim}
\usepackage{booktabs}
\usepackage{siunitx}
\usepackage{graphics}
\usepackage{graphicx,caption}
\usepackage[keeplastbox]{flushend}
\usepackage{suffix}
\usepackage{xstring}
\usepackage{xparse}
\usepackage{expl3}
\usepackage{mathrsfs}
\usepackage{tabularx}
\usepackage{makecell}
\usepackage{array}
\usepackage{hyperref}
\usepackage{cleveref}
\usepackage[ruled,linesnumbered]{algorithm2e}
\usepackage{multirow}
\usepackage{diagbox}
\usepackage{rotating}

\usepackage{subcaption}
\usepackage{wrapfig}

\usepackage{tikz}
\usetikzlibrary{arrows,backgrounds,calc}
\usepackage{relsize}
\usepackage{float}
\usepackage{kantlipsum} 
\usepackage{lipsum}
\usepackage{stfloats}
\usepackage{siunitx}
\usepackage[noadjust]{cite}

\usepackage{todonotes}
\usepackage{soul}
\definecolor{smoothgreen}{rgb}{0.7,1,0.7}
\sethlcolor{smoothgreen}

\newcommand{\Lobject}{\mathcal{L}_{o}}
\newcommand{\Ltarget}{\mathcal{L}_{t}}
\newcommand{\Lsound}{\mathcal{L}_{s}}

\RequirePackage{luatex85}
\usepackage{pgfplots}
\pgfplotsset{compat=newest}
\pgfplotsset{every axis legend/.append style={%
		cells={anchor=west}}
}
\usepgfplotslibrary{polar}
\usetikzlibrary{arrows}
\tikzset{>=stealth'}

\definecolor{C1}{rgb}{0.0, 0.447, 0.741}
\definecolor{C1_light}{rgb}{0.0, 0.6032388663967612, 1.0}
\definecolor{C2}{rgb}{0.85, 0.325, 0.098}
\definecolor{C3}{rgb}{0.929, 0.694, 0.125}
\definecolor{C4}{rgb}{0.494, 0.184, 0.556}
\definecolor{C5}{rgb}{0.466, 0.674, 0.188}
\definecolor{C6}{rgb}{0.301, 0.745, 0.933}
\definecolor{C7}{rgb}{0.635, 0.078, 0.184}

\usepgfplotslibrary{groupplots}

\usetikzlibrary{shapes.geometric, arrows}

\tikzstyle{startstop} = [rectangle, rounded corners, minimum width=2cm, minimum height=1cm,text centered, draw=black, fill=none]
\tikzstyle{arrow} = [thick,->,>=stealth]

\title{
Robot Sound Interpretation: \\ Combining Sight and Sound in Learning-Based Control
}

\author{Peixin Chang, Shuijing Liu, Haonan Chen, and Katherine Driggs-Campbell
\thanks{P. Chang, S. Liu, and K. Driggs-Campbell are with the Department of  Electrical and Computer Engineering at the University of Illinois at Urbana-Champaign. emails: \{pchang17,sliu105,krdc\}@illinois.edu}%
\thanks{H. Chen is with the Computer Engineering Department at Zhejiang University-University of Illinois at Urbana-Champaign Institute. email: haonan2@illinois.edu}%
}

\begin{document}

\maketitle
\thispagestyle{empty}
\pagestyle{empty}

\begin{abstract}

We explore the interpretation of sound for robot decision making, inspired by human speech comprehension. While previous methods separate sound processing unit and robot controller, we propose an end-to-end deep neural network which directly interprets sound commands for visual-based decision making. The network is trained using reinforcement learning with auxiliary losses on the sight and sound networks. We demonstrate our approach on two robots, a TurtleBot3 and a Kuka-IIWA arm, which hear a command word, identify the associated target object, and perform precise control to reach the target. For both robots, we show the effectiveness of our network in generalization to sound types and robotic tasks empirically. We successfully transfer the policy learned in simulator to a real-world TurtleBot3.

\end{abstract}


\section{Introduction}
\label{sec:intro}

People interpret sounds they hear and interact with the world according to their interpretations. When a human maps a sound to meaning, the brain forms conceptual representations of the sound~\cite{kocagoncu2017decoding,hickok2000towards,zhuang2014optimally}. For example, when a person hears the command ``pick up the cellphone,'' her brain associates the sound to the concept of a cellphone so that she can act accordingly. Inspired by human speech comprehension, we explore whether sound commands can be directly interpreted by the robots for visual-based decision making rather than being first transcribed into text.

Traditional voice-controlled robots designs take a different approach. The pipeline first transcribes sound to text with a general-purpose automatic speech recognition (ASR) system. Then, the text is grounded or linked to the physical objects or logic by laborious hard-coding of linguistic and physical rules. The separation of sound interpretation from robotic environments may cause problems when the sound is transcribed without context. For example, a house service robot may mistakenly transcribe the command ``move to the fridge'' as ``move to the feet,'' although the latter is clearly out of context~\cite{vanzo2016robust}. 
Moreover, the language understanding components of the pipeline are mostly rule-based and therefore do not generalize or scale beyond their programmed domains \cite{hermann2017grounded}.

In addition, traditional voice-controlled robots do not fully utilize the information of the sound signals which can reveal tones or even emotions of speakers. This subtle information is lost when a speech is transcribed to text, which may affect the robot's future decision making. Furthermore, some special sounds are ignored by general-purpose ASR systems. Examples include environmental sounds such as doorbells and dog barking. If we want a house service robot to open the door when it hears a doorbell or someone saying ``open the door,'' the traditional ASR powered robots would likely to fail because it cannot transcribe the sounds of doorbells without explicitly training on this data.

\begin{figure}
    \centering
    \includegraphics[scale=0.85]{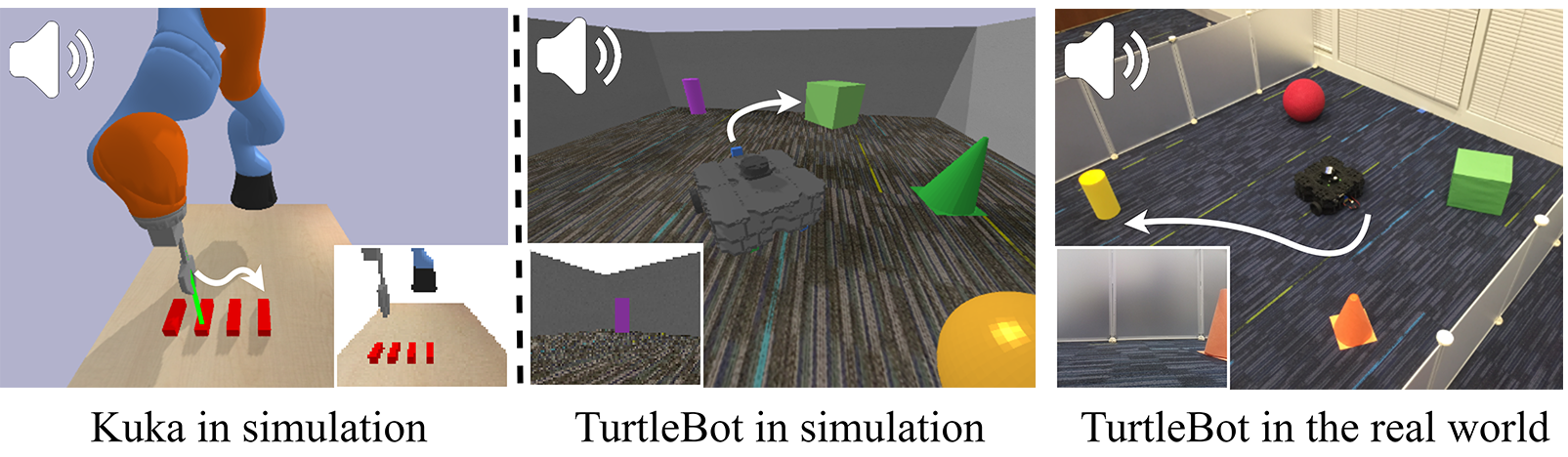}
    \caption{\textbf{Illustration of our robotic tasks.} Windowed image shows the robot camera view. \textit{Left:} The Kuka moves its gripper tip to the target block chosen from four identical blocks. \textit{Middle:} TurtleBot searches for, then navigates to the target object among four objects. \textit{Right:} Experimental setup showing transfer of learned policy to a real-world system.}
    \label{fig:opening}
    \vspace{-15pt}
\end{figure}

In this paper, we seek to create an end-to-end learning framework so that the robots can develop its own interpretations of sound commands for decision making. 
Psychological evidence shows that young children build their conceptual representations through multi-modal embodied sensorimotor activities \cite{smith2005development, wellsby2014developing, meltzoff1990towards}. Similar to children's early development, our framework combines representations from the physical state, vision, and sound. This multi-modal learning experience enables the robot to learn and build its own sound interpretation while interacting with the world, which we call \emph{Robot Sound Interpretation}. Consider the typical scenarios of a voice-controlled robot in Fig.~\ref{fig:opening}. Given a sound command, the robot must identify and approach the corresponding target from the set of objects. To achieve this goal, the robots must learn the meaning of the sound command, draw the correspondence between the visual and audio inputs, and develop a policy to search and reach the target. 

We believe that it is beneficial for robots to develop their own sound interpretation for the following reasons. First, the sound feature representations can be used directly for communication without being first translated into text for decision making, which reduces accumulative errors. Second, instead of separating sound transcriber from the context, the robots learn the meaning of sounds while interacting with its environment. Thus, the grounding process is simplified and the learned representation is tightly related to the task without out-of-context problems. Third, our system is general in terms of types of sounds, robots, and robotic tasks, and the learned sound interpretations can change and evolve.

The main contributions of this paper are: (1) We present the novel concept of Robot Sound Interpretation---learning policies directly from sound commands without translating sound to text. (2) We develop a network architecture that is trained end-to-end without hand-crafted features, allowing us to use a wide variety of sound inputs on different robot environments to customize the robot's interpretation of sounds. (3) Our network exhibits promising results in two robot platforms: a Kuka-IIWA and TurtleBot3 in simulation. The TurtleBot3 policy was successfully transferred to a real-world TurtleBot3 without training on real-world data.

This paper is organized as follows: 

In Section~\ref{sec:methods}, we formalize the problem and propose our network architecture. Experiments in simulation and in the real-world are discussed in Section~\ref{sec:exp} and Section~\ref{sec:real_exp}, respectively. Finally, we conclude the paper in Section~\ref{sec:conclusion}.
\section{Related Works}
\label{sec:related}
\subsection{Voice-Controlled robots}

Motivated by human's spoken language communication, many robotics researchers have investigated voice-controlled robots. The usual pipeline consists of three stages. First, the speech signal is processed by an automatic speech recognition (ASR) system, which transforms speech to text~\cite{cruz2016multi,fernandez2016natural,burger2012two}. Second, the text is analyzed and linked to the logic relations or entities in the physical world~\cite{bastianelli2016discriminative}. The meaning of the instruction is interpreted and understood by the robot in this stage. Third, the robot plans and performs actions based on the analysis~\cite{stramandinoli2016grounding,paul2018efficient,ovchinnikova2015multi,liu2019review}. Although the pipeline can be applied to large corpuses and can handle sentences, the pipeline suffers from several limitations. In the first stage, off-the-shelf ASRs usually operate in a general purpose setting and their outputs could be out-of-context to specific robotic tasks \cite{vanzo2016robust,twiefel2014improving}. Special treatments are needed to select the best candidate from a list of transcriptions produced by the ASRs \cite{cruz2016multi,stramandinoli2016grounding}. The second stage of the pipeline is mostly hard-coded and rule-based. Thus, a small error in any stage can cause the failure of the entire system. However, in our method, since the agent is situated, the meaning of the sounds is understood within the context, and our model is more customizable to variations of sound types and sound to meaning mappings.  

\subsection{Natural language grounding with deep learning}
Inspired by the recent success of deep learning, situated language learning agents have been developed to execute tasks such as navigation \cite{anderson2018vision,chen2019touchdown,yu2018interactive} and object finding \cite{hermann2017grounded,chaplot2018gated} according to natural language instructions and images. 

These works successfully show that a robot may learn to understand language and perform grounding in an end-to-end fashion.
However, compared to our work, these studies are typically conducted in highly controlled simulated environments with simpler motion models and, to our knowledge, these agents have not been transferred to a physical robot platform for performing a real-world task. Moreover, these agents take text-based instructions, not sound signals, and thus abstracting away the complexities of speech and limiting the potential of human-robot communication. In contrast, we use raw acoustic signals as one of the inputs and demonstrate our robot tasks in both simulation and the real world. 

\subsection{Learning audio feedback}
Acoustic signals can be used as a feedback for deep learning based robot control tasks. Liang \textit{et al}. utilized the audio vibration to estimate liquid height in a container in robotic liquid pouring tasks~\cite{liang2019making}. Lathuili\`ere \textit{et al}. utilized audio signals as one of the inputs to control the gaze of a robot~\cite{lathuiliere2019neural}. Although these works showed the benefits of training robots in acoustic environments, the problem setting of our work is completely different. We focus on voice-controlled robots that must understand sound commands correctly at the beginning of robot's task execution. 

\subsection{Cognitive architectures}
Cognitive architectures such as \cite{zhang2001grounded,liu2001robot} also attempt to mimic the process of human speech comprehension. However, the architectures in these works directly map sounds to predefined robot actions in an one-to-one manner, making the agents less capable of performing long-term decision making~\cite{zhang2001grounded}. Further, these works typically only utilize the sound modality, overlooking the interplay between sound and sight~\cite{manoonpong2005neural}. Our model actively learns a complex action sequence combining information from images and audios.

\section{Methodology}
\label{sec:methods}

\subsection{Problem formulation}
Consider an agent interacting with an episodic environment $\mathcal{E}$. We model this interaction as a Markov Decision Process, defined by the tuple $ \langle \mathcal{X}, \mathcal{A}, P, R, \gamma, \mathcal{X}_0 \rangle$. For each episode, the agent begins at a random initial state $x_0\in \mathcal{X}_0$. At each time step $t$, the state $x_t\ = \{S, I_t, M_t\}\in\mathcal{X}$ of the agent consists of three parts: a one-time sound feature ($S$) representing the command, an image ($I_t$) from its camera, and a robot state vector ($M_t$) which includes information such as manipulators' end effector location or locomotors' odometry. 
The agent then takes an action $a_t\in\mathcal{A}$ according to its policy $\pi_\theta(a_t|x_t)$ parameterized by $\theta$. In return, the agent receives a reward $r_t$ and transitions to the next state $x_{t+1}$ according to an unknown state transition $P(\cdot|x_t, a_t)$. 
The process continues until $t$ exceeds the maximum episode length $T$ and the next episode starts. Let $\gamma\in(0,1]$ be the discount factor. Then,  $R_t=\sum^\infty_{k=0}\gamma^{k}r_{t+k}$ is the total accumulated return from time step $t$. The goal of the agent is to maximize the expected return from each state. The value of state $x$ under policy $\pi$, defined as $V^{\pi}(x)=\mathbb{E}[R_t|x_t=x]$, is the expected return for following policy $\pi$ from state $x$. 

\begin{figure}[t!]
    \centering
    \includegraphics[scale=0.5]{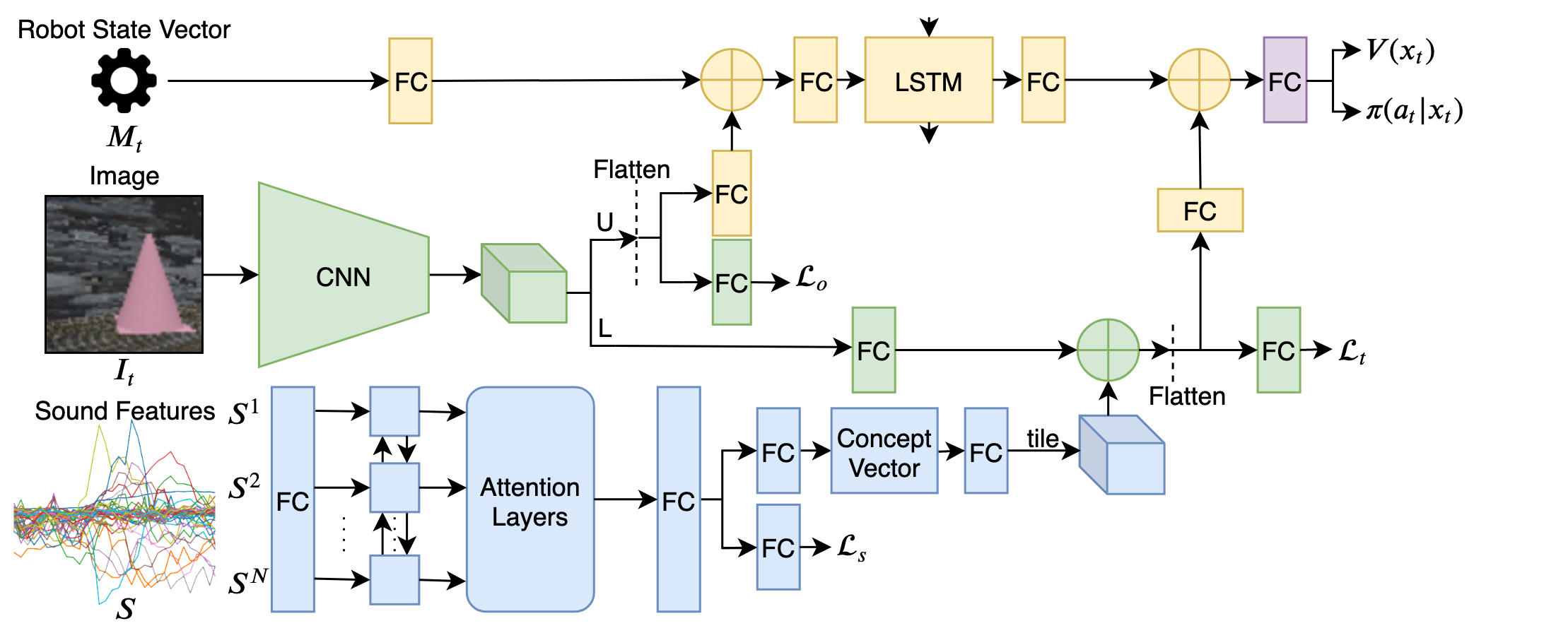}
    \caption{\textbf{The network architecture}. The sound interpreter is blue, the visual-motor integrator is yellow and green, and the policy learner is purple. 
    The losses $\mathcal{L}_{o}$ and $\mathcal{L}_{t}$ are only needed in the more challenging TurtleBot environment, while $\mathcal{L}_{s}$ is applied in both environments. We use $\bigoplus$ to denote element-wise addition and $FC$ to denote fully connected layers.}
    \label{fig:wholeNet}
    \vspace{-20pt}
\end{figure}

\subsection{Network Architecture}
The proposed network, as shown in Figure~\ref{fig:wholeNet}, is comprised of three interactive components: the sound interpreter, the visual-motor integrator, and the policy learner. 

\subsubsection{The sound interpreter}
The sound interpreter takes in sound feature $S$ and outputs an encoding, which we call the concept vector $C$. This vector is the agent's own interpretation of the sound command, which is formed through its interaction with the environment. This vector encapsulates the information for the sound and mimics the functionality of humans’ conceptual representations when they map sound to meaning. The sound feature $S\in\mathbb{R}^{N \times m}$ is the Mel Frequency Cepstral Coefficients (MFCCs) of the raw acoustic signal \cite{davis1980comparison}, where $N$ is the number of frames of the signal and the $m$ is the number of MFCCs for a frame. We use $S^j$ to denote the $j^{th}$ frame of $S$, where $j\in\{1,2,...,N\}$. 

A single layer bi-directional long short-term memory (BiLSTM) network with number of units $d_S$ and zero initial state is used to encode $S$ \cite{schuster1997bidirectional}. Let $h_j=[\overrightarrow{h}_j^\top, \overleftarrow{h}_j^\top]^\top \in\mathbb{R}^{2d_S}$ be the concatenated output of the forward and backward cells of the BiLSTM at frame $j$. The BiLSTM output for $S$ is $h=[h_1^\top; h_2^\top;...;h_N^\top] \in \mathbb{R}^{N \times 2d_S}$. An attention layer over the BiLSTM is applied to select the relevant temporal locations over the input sequence. The final hidden states $h_N$ is first expanded to a $N \times 2d_S$ matrix. Let $h_A\in \mathbb{R}^{N \times 4d_S}$ be the result of concatenation between $h$ and the expanded $h_N$. The unnormalized score $e\in\mathbb{R}^{N}$ for $h$ is $e=f(h_A) v_A$, where $f(\cdot)$ are fully connected layers, the last of which has $l$ units, and $v_A\in\mathbb{R}^{l}$ are trainable weights. The weight $\alpha\in\mathbb{R}^{N}$ for each $h_j$ in $h$ is calculated by a softmax function: $\alpha=\text{softmax}(e)$. The weighted sum of these $h_j$, denoted $c\in\mathbb{R}^{2d_S}$, is computed by multiplying the concatenated output of the BiLSTM with the weight $\alpha\in\mathbb{R}^{N}$: $c=h^\top\alpha$. The concept vector is given by $C=g(c)$, where $g(\cdot)$ are fully connected layers with $\text{tanh}$ activation that transforms $c$ into a 64-dimensional vector. 

Compared to the attention mechanism used in the literature of neural machine translation and automatic speech recognition \cite{chan2015listen,bahdanau2014neural,luong2015effective}, our attention mechanism uses a score function that does not depend on a decoder state, as our model does not explicitly have a decoder.  
We found multi-head attention beneficial in our tasks~\cite{vaswani2017attention}. 
In multi-head attention, each head generates a different attention distribution, which allows the model to learn to attend to different locations. The agent using multi-head attention usually achieves a slightly higher average reward. 

At test time, the sound interpreter only encodes $S$ at the first time step. The resulting concept vector, $C$, is cached and shared throughout an episode. There are two reasons for first-step encoding. First, first-step encoding mimics how we humans perceive and respond to sound commands. While the sound is transient, the interpretation of the sound can be temporarily stored and then used to finish the intended tasks after the sound has diminished. Second, first-step encoding makes our system more efficient at test time because the sound interpreter only needs to perform inference at the first time step instead of every time step. The delay between two neighboring time steps caused by the inference can be decreased. 

We add an auxiliary adaptation loss $\mathcal{L}_{s}$ in the sound interpreter, which predicts the label of the received acoustic signal. The $\mathcal{L}_{s}$ is a multi-class classification loss which helps the sound interpreter for feature extractions. 

\subsubsection{Visual-motor integrator}
The visual-motor integrator takes the concept vector $C$ produced by the sound interpreter, the current visual observation $I_t$, and the current robot state vector $M_t$ as the inputs and creates a joint representation which will be used by the policy learner. The image $I_t$ is processed by a VGG-style convolution neural network (CNN) \cite{simonyan2014very}, the output of which is shared by two branches denoted by $U$ and $L$ in Fig. \ref{fig:wholeNet}. The upper branch $U$ merges with the features extracted from the robot state vector, and the resulted vector is fed into a single-layer LSTM, whose output is responsible for agent's visual-motor skill. The LSTM allows the model to capture information from the past, which is important for a partially observable environment. The lower branch $L$ merges with the information from the concept vector $C$, which forms a joint representation between images and the agent's interpretation of the sound. 

The visual-motor integrator includes two task specific auxiliary losses, $\mathcal{L}_{o}$ and $\mathcal{L}_{t}$. The loss $\mathcal{L}_{o}$ is a multi-label classification loss that predicts which of the four objects are in $I_t$, providing a guidance of visual feature extraction for object recognition. The loss $\mathcal{L}_{t}$ is a binary classification loss that predicts whether the target object is in $I_t$, and thus builds an association between the command heard by the agent and the corresponding object in the arena.  
 
\subsubsection{Policy learner}
The output of the visual-motor integrator is fed into the policy learner, whose loss function is denoted as $\mathcal{L}_{pg}$.  We use Proximal Policy Optimization (PPO), a model-free policy gradient algorithm, for policy and value function learning \cite{schulman2017proximal}. We adopt the implementation from OpenAI Baselines and modify the original implementation so as to incorporate various types of observation from the environment \cite{baselines}. Eight instances of the environment run in parallel for collecting the agent's experiences, which both accelerates and stabilizes learning. When updating the policy, the entire history of four episodes is used for an update. 

\subsubsection{Joint architecture}
 
The whole architecture combines all components above. The whole network can be trained end-to-end and the agent's interpretation of sounds is optimized by the gradient from all other components of the network. In this way, our embodied agent learns the meanings of sound commands from its own sensory-motor experience instead of completely from ground truth labels.

Our loss function $\mathcal{L}_{tot}$ is a linear combination of $\mathcal{L}_{pg}$, $\mathcal{L}_{s}$, $\mathcal{L}_{o}$, and $\mathcal{L}_{t}$, giving: 
\begin{equation}
   \mathcal{L}_{tot}=w_{pg}\mathcal{L}_{pg} + w_{s}\mathcal{L}_{s} + w_{o}\mathcal{L}_{o} + w_{t}\mathcal{L}_{t}
\end{equation}
where $w_{pg}$, $w_{s}$, $w_{o}$, and $w_{t}$ are the scalar weights for each loss. We set $w_{pg}=w_{s}=1$ for both environments, $w_{o}=w_{t}=0$ for the Kuka environment, and  $w_{o}=w_{t}=0.5$ for the TurtleBot environment. The learning rate is $8e^{-6}$. 

\section{Simulation experiments} 
\label{sec:exp}
In this section, we describe the sound data processing and present two test cases in two robotic environments, as well as training and testing the models in simulation.
$\vspace{-5pt}$
\subsection{Sound data processing}
To show that our model is capable of handling various types of sounds, we consider three types of acoustic signals: single-word speech signals, environmental sounds, and single-tone signals. 

For speech signals, we choose two sets of words from the Speech Commands Dataset~\cite{warden2018speech}. Wordset1 contains the utterances of four numbers: ``zero,'' ``one,'' ``two,'' and ``three.'' Wordset2 includes the utterances of four objects: ``house,'' ``tree,'' ``bird,'' and ``dog.'' In the data preprocessing stage, we clean the datasets by removing highly distorted and blank recordings. For each word in the two sets, 1,000 samples are used for training and 50 samples are used for testing. 

For environmental sounds, we use UrbanSound8K Dataset \cite{Salamon:UrbanSound:ACMMM:14}. We choose ``car horn,''  ``gun shot,'' ``dog bark,'' and ``jackhammer''.\footnote{For ``car horn,'' we use both foreground and background sound because the amount of data in this class is small. For all the other classes, we use only foreground sound.} For each class, at least 400 samples are used for training and 50 samples are used for testing. 

For single-tone signals, we choose four musical notes, $C_4$ (Middle C), $D_4$, $E_4$, and $F_4$, performed by various musical instruments from NSynth datasets \cite{nsynth2017}.
For each of the four musical notes, 1,000 samples are used for training and 50 samples are used for testing. 

When extracting the MFCCs for the sound signals, we use 40 filters in the filter bank and 40 coefficients~\cite{pythonSpeechFeatures}. The resulted MFCCs is padded with zeros if the number of frames is less than 100 so that the sound feature $S\in\mathbb{R}^{100 \times 40}$.

\subsection{Robotic Environments}
We design two environments to show that our model can be applied to different robotic platforms and tasks. The environments are implemented using PyBullet~\cite{coumans2019}. The Kuka environment uses a location-fixed Kuka-IIWA arm, while the TurtleBot environment uses a mobile TurtleBot3. 
 
\subsubsection{TurtleBot} The TurtleBot environment consists of a $2$m $\times$ $2$m arena  and
four fix-sized 3D objects: a cube, a sphere, a cone, and a cylinder. The action $a=[\delta_{v}, \delta_{a}]^\top$, where $\delta_{v}$ is the change of the desired transitional velocity $v_{d}$ in $m/s$, and $\delta_{a}$ is the change of the desired angle $\phi_{d}$ in $rad$ with respect to robot's initial orientation. We define the update rule for $v_d$ and $\phi_d$ with respect to time $t$ as  
\begin{equation}
\begin{split}
    v_{d}[t] &=0.05\delta_{v}+v_{d}[t-1]
    \\
    \phi_{d}[t] & =0.15\delta_{a}+\phi_{d}[t-1]
\end{split}
\end{equation}
We apply a simple proportional control on the desired angle.

We match each object with a sound. When an episode begins, the robot is placed randomly near the center of the arena. The four objects with randomly chosen color are placed randomly around the robot. The environment then randomly chooses one sound from one of the four words or musical notes. The MFCCs of the sound, $S$, are extracted. At every timestep, the TurtleBot receives a $75\times100\times3$ RGB image from its camera, which is cropped and resized to $I_t$ of size $96\times 96\times 3$ for CNN processing. The robot state vector $M_t=\left[v_{d}[t],\phi_{d}[t]\right]$. The time horizon, $T$, is 80. We define the reward $r_t=r_{c}+r_{d}+r_{a}+r_{goal}$. Let $d_t$ and $\beta_t$ be the safe distance and angle from the camera's central axis to the target object at time $t$. $r_{d}$ and $r_a$ are directly proportional to the robot's progress in distance and angle, respectively. The collision penalty, $r_c$, is $-0.1$ if the robot is too close to the target object and $-0.3$ if the robot is too close to the other objects and the wall. 
\begin{equation}
\begin{split}
\begin{gathered}
	r_{d} = 50(-|d_t|+|d_{t-1}|) \\[-3pt]
    r_{a}= 20(-|\beta_t|+|\beta_{t-1}|) 
\end{gathered}
\end{split}
\end{equation}
\vspace{-5pt}
\begin{equation*}
\begin{split}
\begin{gathered}
    r_{goal}  = 
        \begin{cases}
            2, & \text{if } z_d^-\leq d_t \leq z_d^+ \text{ and } z_{\beta}^-\leq |\beta_t| \leq z_{\beta}^+\\
            0.5, & \text{if } z_d^-\leq d_t \leq z_d^+ 
            \text{ or } z_{\beta}^-\leq |\beta_t| \leq z_{\beta}^+\\
            0, & \text{otherwise}.
        \end{cases}
\end{gathered}
\end{split}
\end{equation*}
where $z_d^+$ and $z_d^-$ are the upper and lower limits on the goal distance, and $z_\beta^+$ and $z_\beta^-$ are the upper and lower limits on the goal angle. Intuitively, the agent gets high reward when it approaches to the target, while maintaining a safe distance. 

\subsubsection{Kuka} The Kuka environment consists of a table, a Kuka-IIWA arm, and four identical blocks with length 0.1m and width 0.04m. The blocks are placed in a line parallel to one side of the table in front of the robot with even spacing. The gripper tip can only move within an xy-plane, above the table top. The action $a=[\delta_{x}, \delta_{y}]^\top$, where $\delta_{x}$ is the change of the desired gripper tip location in the $x$ axis $p_{x}$,  and $\delta_{y}$ is the change of that in $y$ axis $p_{y}$ at time $t$.  We define the update rule for $p_{x}$ and $p_{y}$ with respect to time $t$ as
\begin{equation}
\begin{split}
    p_{x}[t]=0.01\delta_{x}+p_{x}[t-1]\\
    p_{y}[t]=0.01\delta_{y}+p_{y}[t-1]
\end{split}
\end{equation}
Both $p_{x}$ and $p_{y}$ are limited within a certain range where the gripper tip can move. We use position control for controlling the joints of the robot. 

We match each block with one word or note. When an episode begins, the location of the gripper tip and the blocks are initialized randomly and the environment randomly chooses one sound from one of the four words. The MFCCs of the sound, $S$, are extracted. At every timestep, the agent receives an $80\times 80\times 1$ black and white image $I_t$ from a fixed RGB camera. From the image, the robot can always see the four blocks and its gripper tip. The robot state vector $M_t=\left[l_{x}[t],l_{y}[t]\right]^\top$, where $l_{x}[t]$ and $l_{y}[t]$ are the gripper tip location at time $t$. The time horizon $T$ is 200. We define the reward $r_t=r_{d}+r_{goal}$. Let $d_t$ be the distance between the robot gripper tip and the target block center at time $t$ in the xy-plane, then $r_{d}$ and $r_{goal}$ is defined as 
 \begin{equation}
\begin{split}
     r_{d} &= -0.5d_t - 0.1 (\ln{(d_t^2 + 5 \times 10^{-5})} + 3.5),
    \\
   	r_{goal} &= 
    \begin{cases}
    1, & \text{if } M_t\in A\\
    0, & \text{otherwise}.
    \end{cases}
\end{split}
\end{equation}
where $A$ is the area right above the target block. Intuitively, the agent gets high reward when its gripper is at the center of the target block.

Agents trained in these environments will encounter different difficulties. The agent in the Kuka environment needs to develop spatial reasoning skills that can differentiate the target object from the four identical blocks using their relative positional information observed from only one camera. The agent in the TurtleBot environment needs to learn exploration skills so as to find the target object as quickly as possible.

 \subsection{Experiment setup}

We now introduce the training procedure and evaluation of our network in simulation. In our experiments, the overall architecture of the network is kept the same in both TurtleBot and Kuka environments, except the following modifications. 

In TurtleBot environment, we pretrain the vision module (green component in Fig.~\ref{fig:wholeNet}) to accelerate the convergence of the whole network. We collect image data through a self-supervising process for all sixteen labels in simulator to optimize $\Lobject$, where each label indicates whether an object is in $I_t$. Using the label of these image,  we optimize $\Ltarget$ by replacing the concept vector to randomly generated one-hot vectors. TurtleBot environment requires a deeper CNN and a deeper LSTM to achieve good performance, because the partial observability of this environment adds complexity to the robot's task. We train the network for $1.5\times 10^6$ timesteps.

In Kuka environment, we do not need pretrained weights to obtain a good policy as the task requires less exploration. $\Lobject$ and $\Ltarget$ are no longer needed since all four identical blocks are always in the robot's fixed camera view. We train the network for $5\times 10^6$ timesteps.

We create a Mix dataset to show that our model can map different sounds to one object. Mix dataset combines Wordset2 with UrbanSound8K Dataset. The ``tree'' class and the ``bird'' class are the same as Wordset2; the other two classes contain an even mixture of data from two different sub-classes: ``house'' is mixed with ``jackhammer,'' ``dog'' is mixed with ``dog bark.'' The model must learn that more than one classes of words have the same meaning during training, and therefore the model's concept vector forms an abstract meaning of these objects. For each class, 1,000 samples are used for training and 50 samples are used for testing.

 To show that the agent's interpretation of a sound can evolve dynamically, we conduct an experiment denoted as ``Wordset2 $\rightarrow$ Mix.'' Specifically, the agent first learns in the TurtleBot environments for $1.5\times 10^6$ timesteps or in the Kuka environment for $5\times 10^6$ timesteps with Wordset2. Then, the model is trained for additional timesteps with our Mix dataset. In other words, the agent encounters new sounds of ``jackhammer'' and ``dog bark,'' and must learn that both ``dog'' and ``dog bark'' refer to the same target object. This follows for ``house'' and ``jackhammer.''

\subsubsection{Evaluation criteria}
In simulation, we test the policy for $200$ episodes ($50$ for each target object) with unheard sound data. For TurtleBot environment, we define an episode as success if the robot stays close to the target object for more than $20$ timesteps. For Kuka environment, we define an episode as success if the gripper stays close to the target block for more than $100$ timesteps. We define success rate as the percentage of successful episodes in all test episodes and use it as our metric.

\subsubsection{Baselines} 

We introduce a random walk baseline as the poor standard for comparison, where the robots take a random action at each timestep without any training. 

\begin{figure*}[ht]
    \centering
    
    TurtleBot Environment\\
    \includegraphics[scale=0.29]{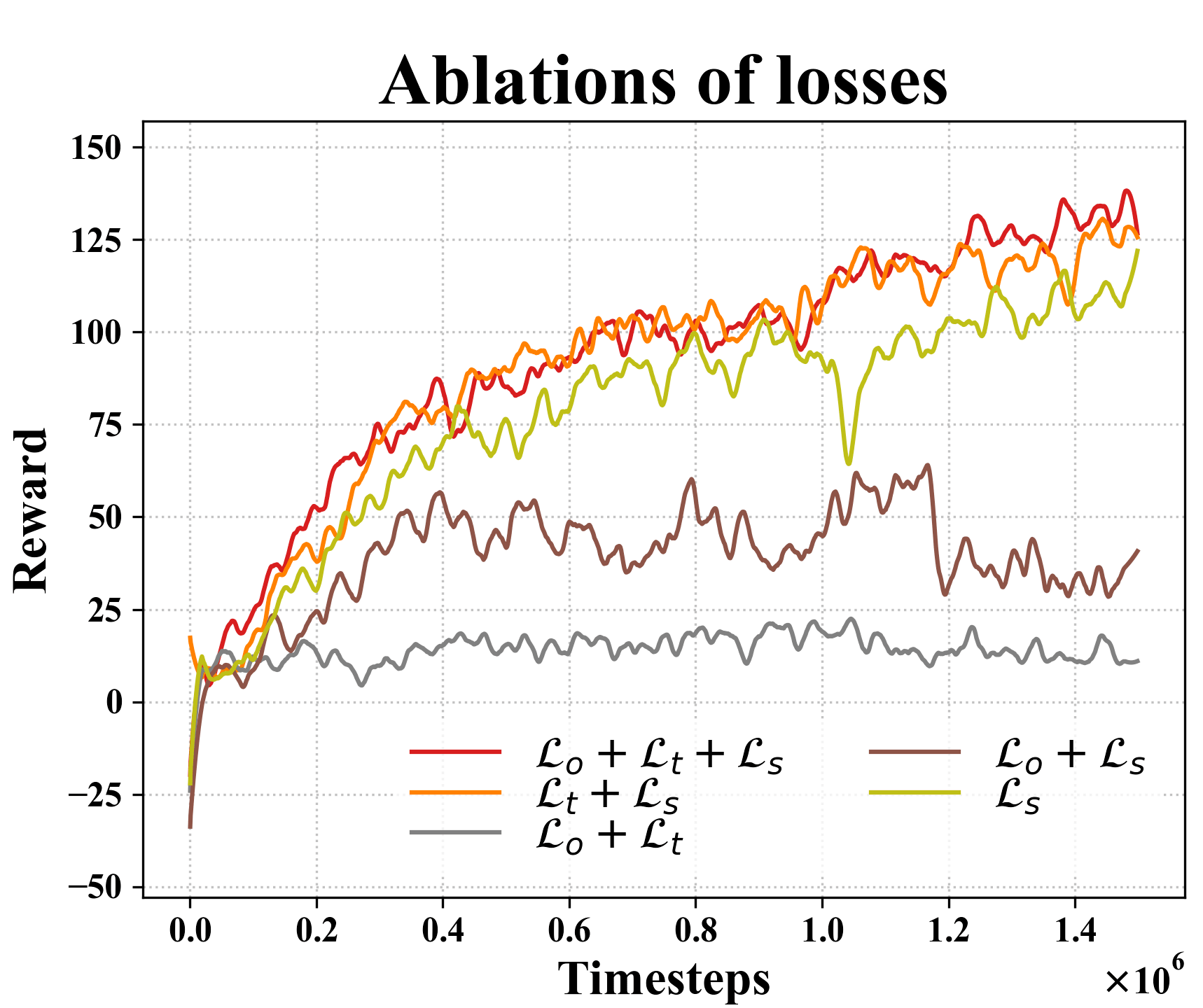}
    \includegraphics[scale=0.29]{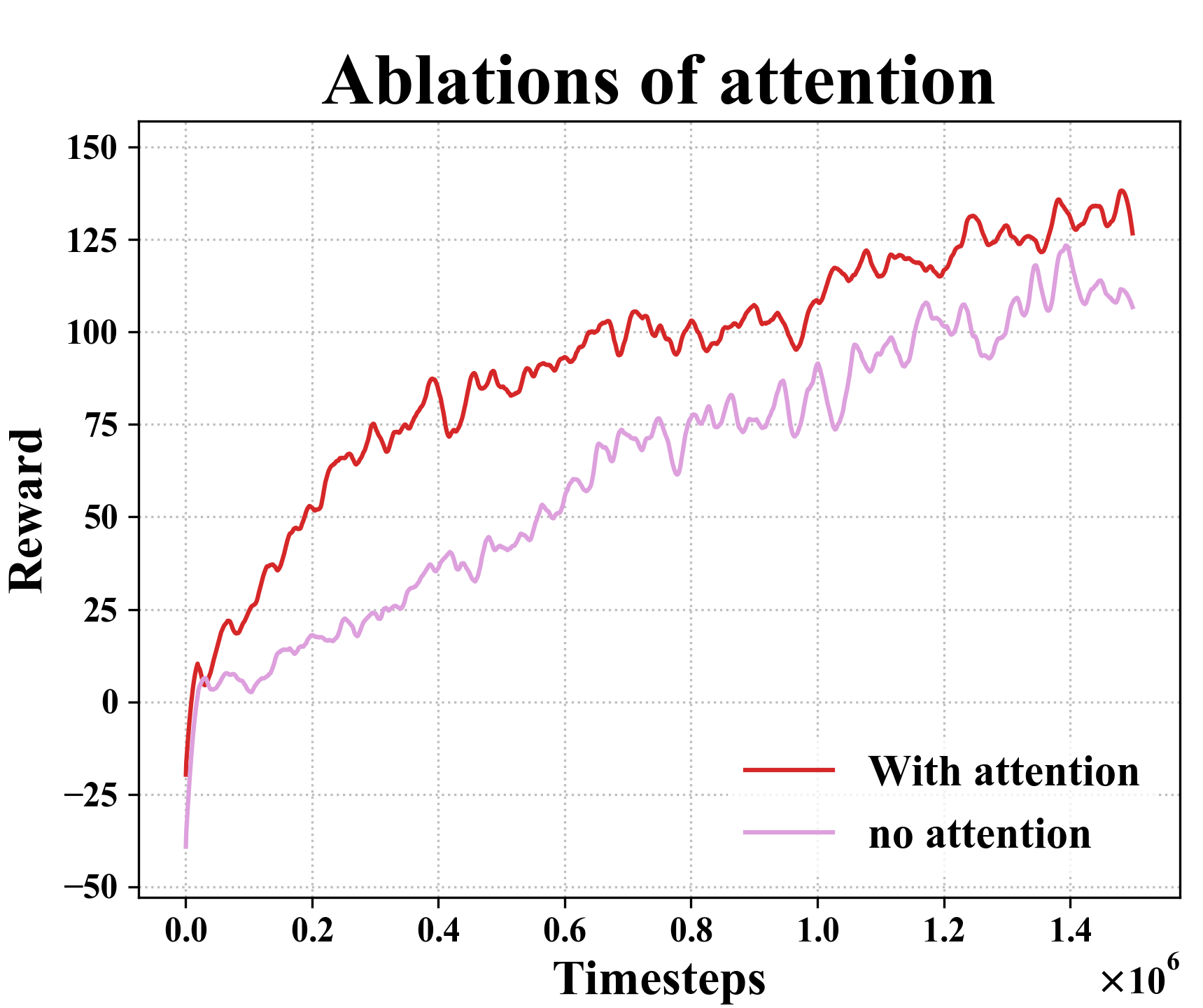}
    \includegraphics[scale=0.29]{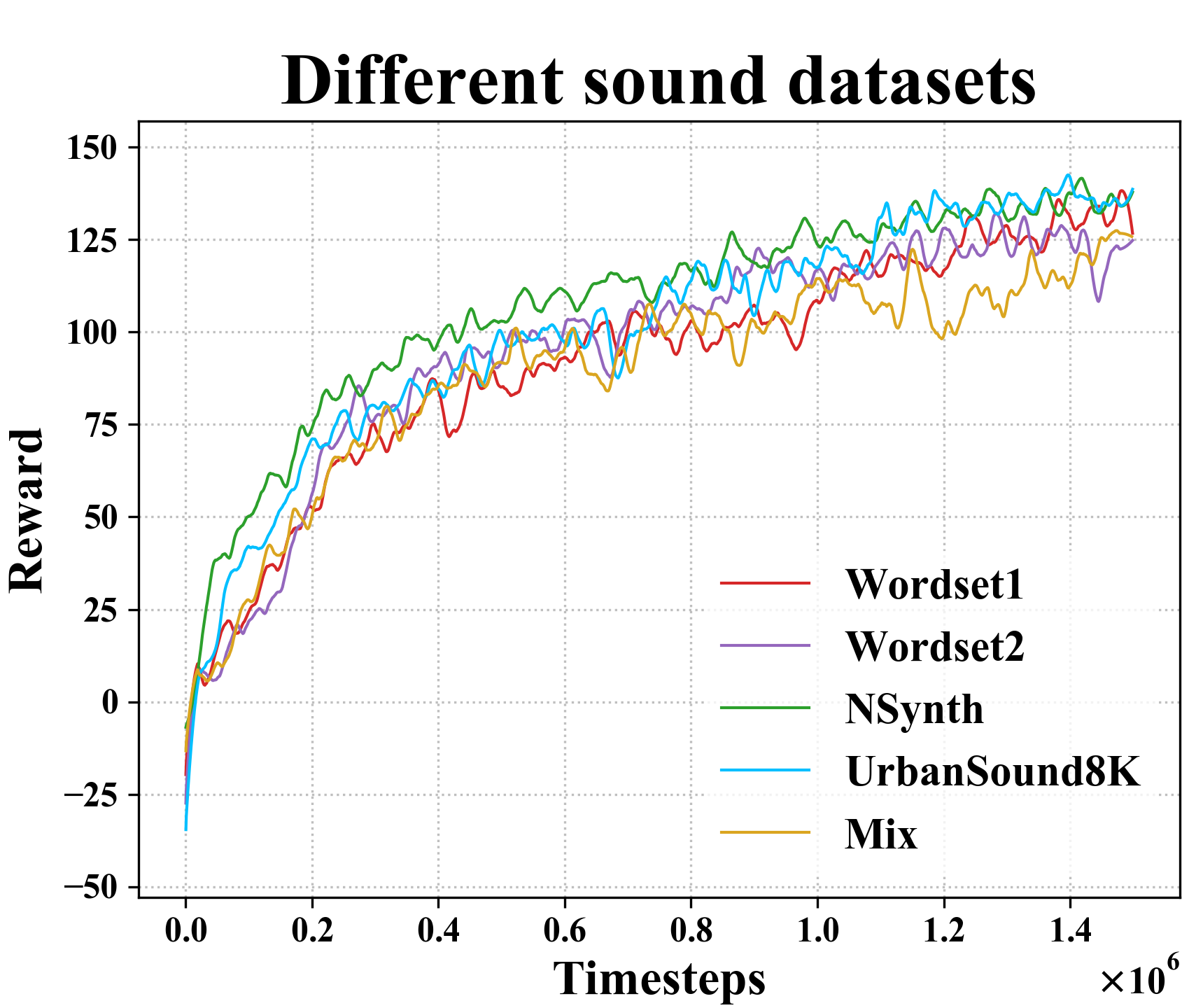}
    \includegraphics[scale=0.29]{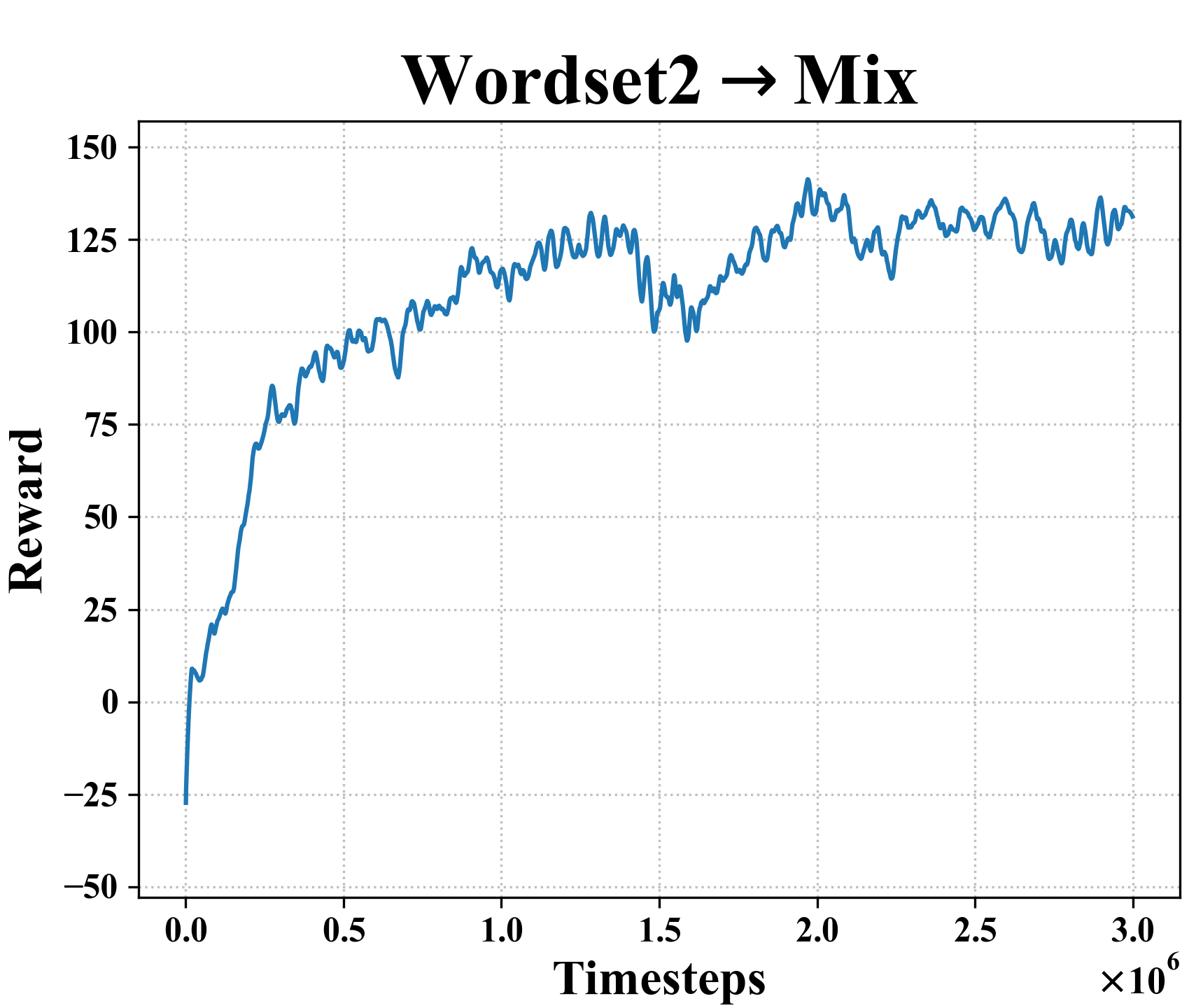}
    
    Kuka Environment\\
    \includegraphics[scale=0.29]{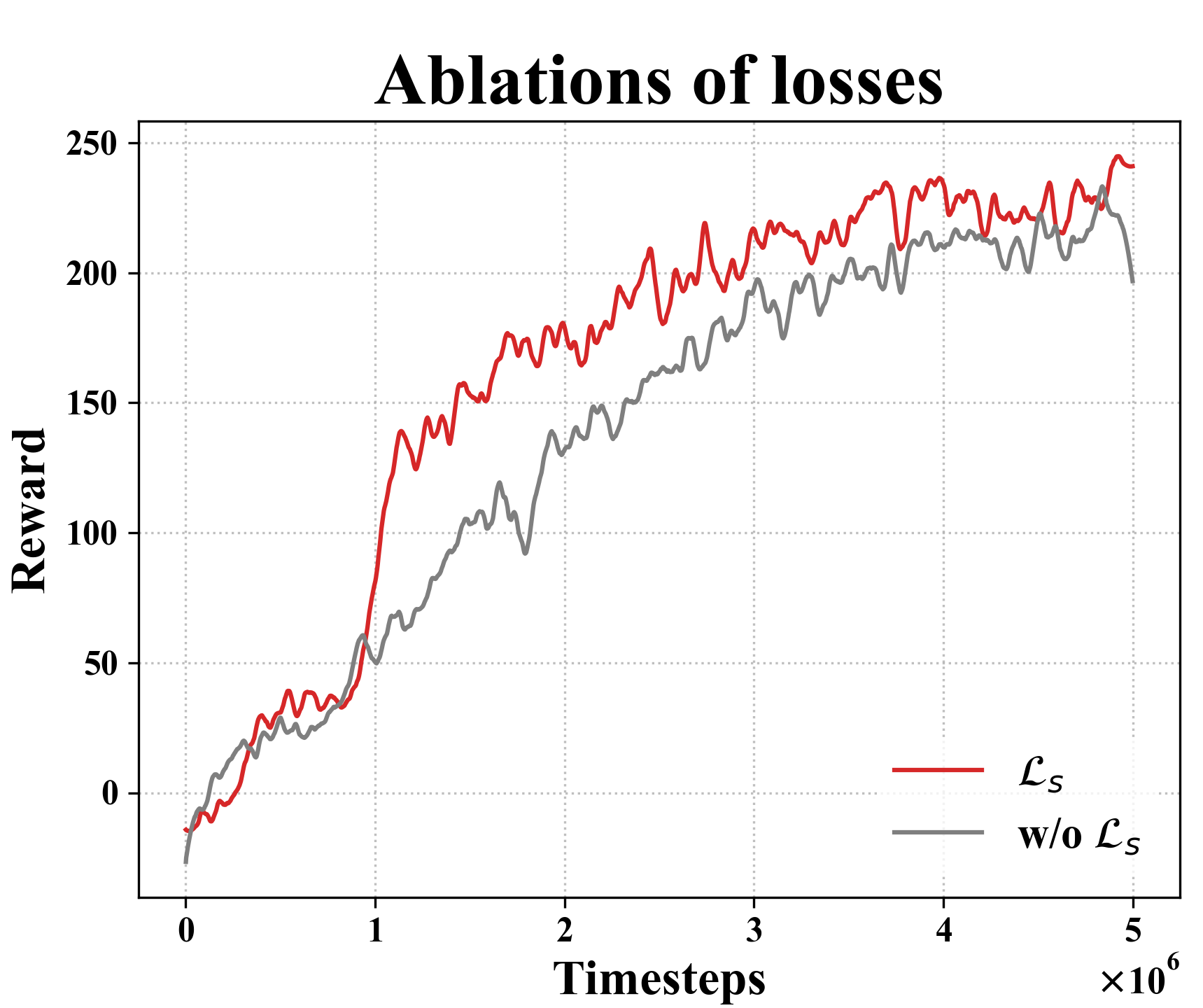}
    \includegraphics[scale=0.29]{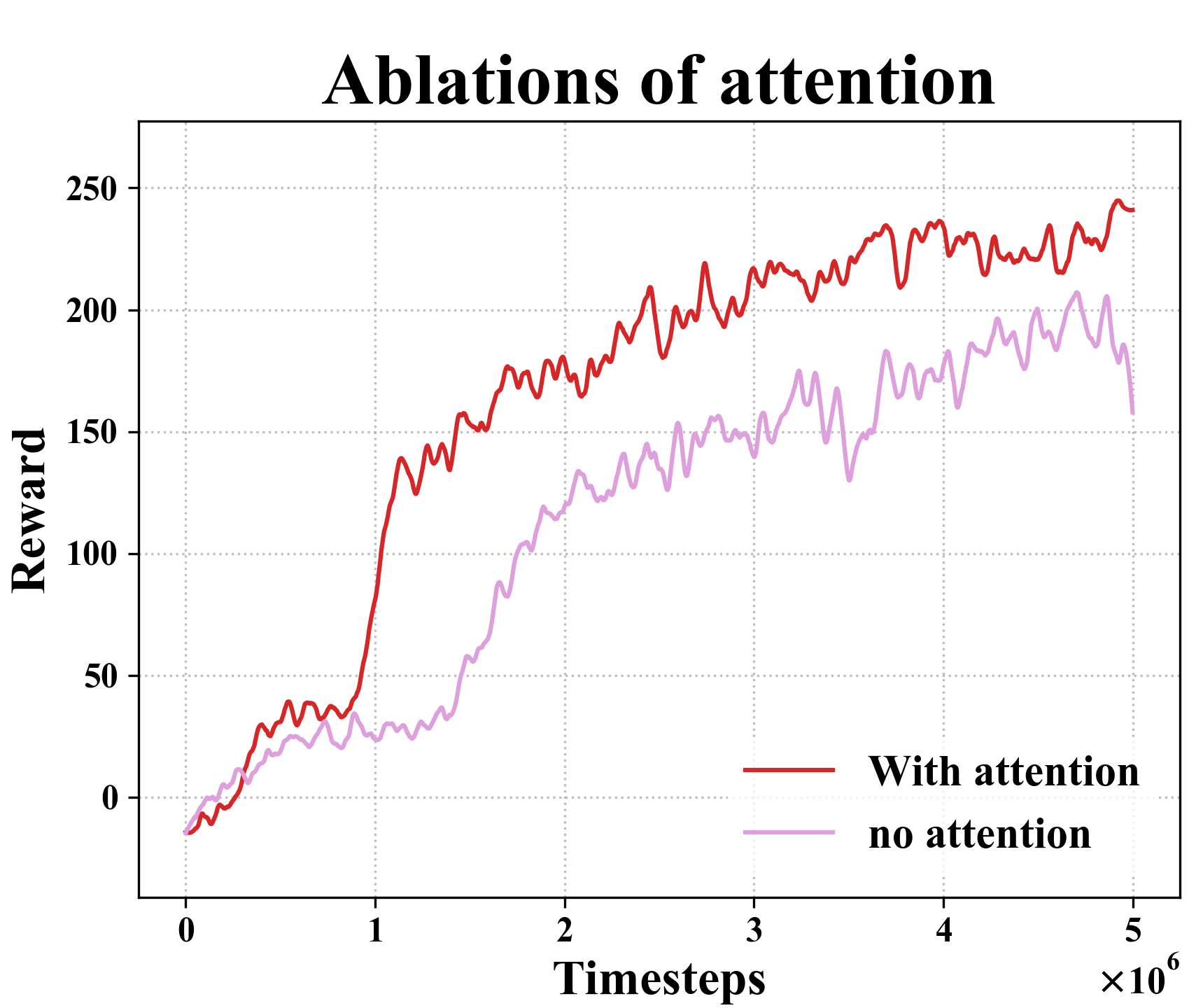}
    \includegraphics[scale=0.29]{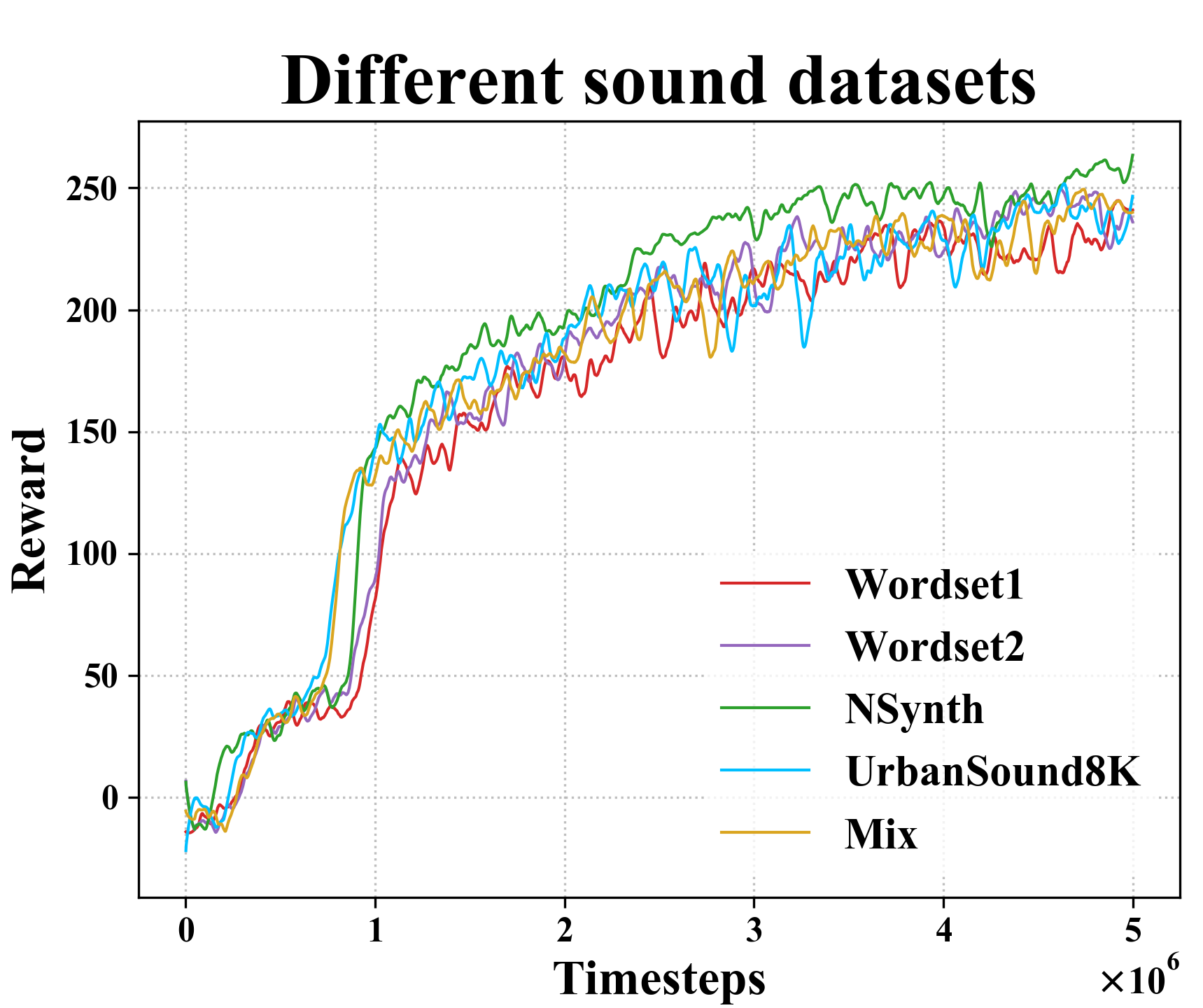}
    \includegraphics[scale=0.29]{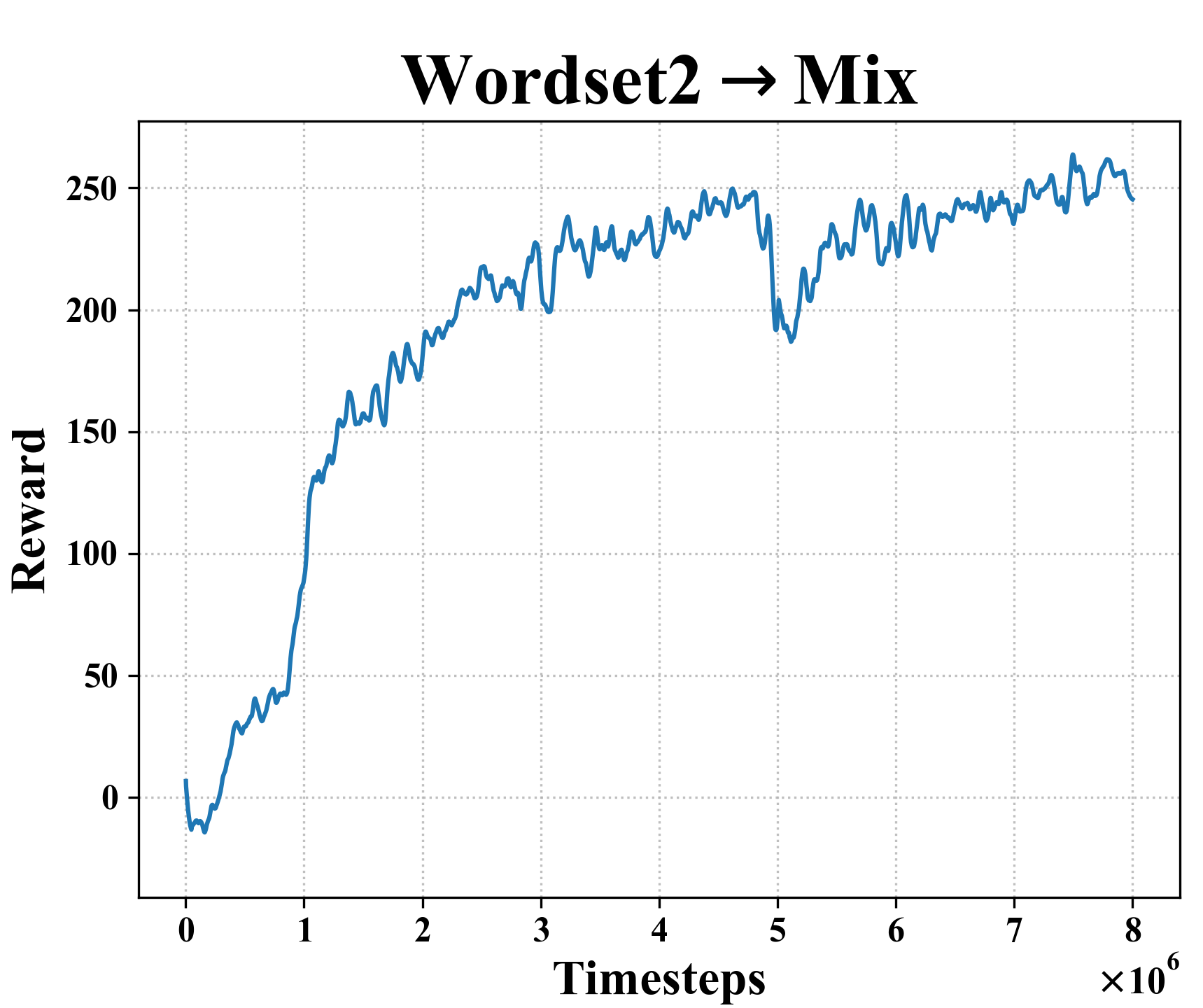}
    \caption{\textbf{Training curves of our network in TurtleBot environment (top) and Kuka environment (bottom).} \emph{Left:} Comparison of our trained model and models without different kinds of auxiliary losses. All models have 4 attention heads and are trained with Wordset1. \emph{Middle Left:} Comparison of our model with and without attention mechanism, both of which have all three auxiliary losses and are trained with Wordset1. \emph{Middle Right:} Comparison of our model trained with different sound data, all of which have all three auxiliary losses and 4 attention heads. \emph{Right:} Learning process for the experiment ``Wordset2 $\rightarrow$ Mix''}
    \label{fig:turtlebot_result}
    \vspace{-15pt}
\end{figure*}

\textbf{}\begin{figure}
    \centering
    \includegraphics[scale=0.17]{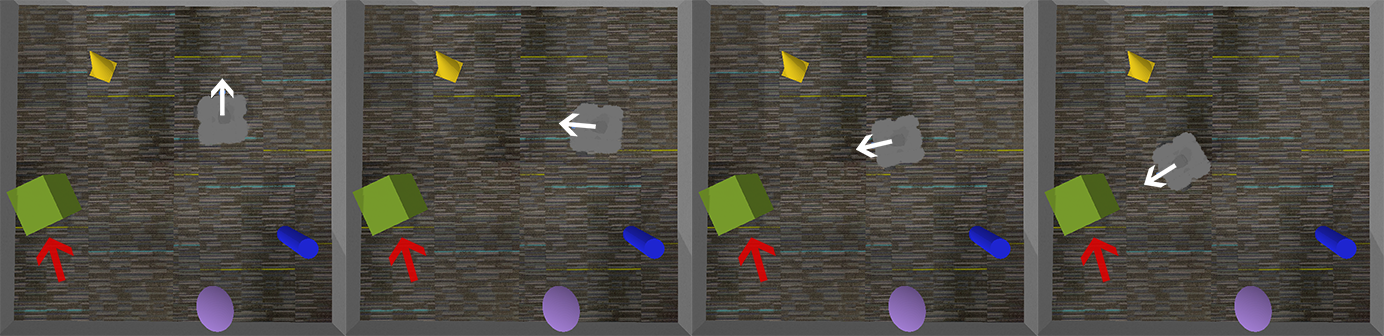}
    \includegraphics[scale=0.17]{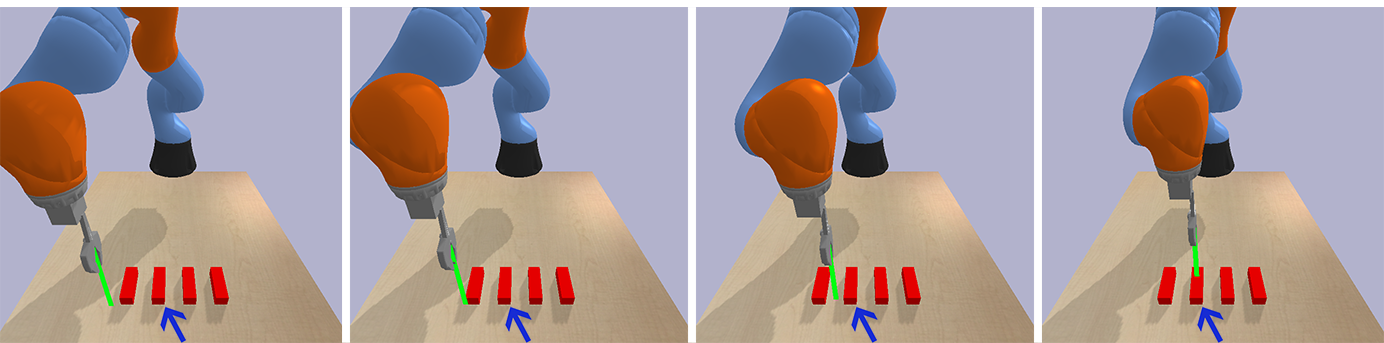}
    \caption{\textbf{Trained model in simulation}. \textit{Top}: Simulated TurtleBot searches and approaches its target, the green cube. The heading direction is shown by white arrows. \textit{Bottom}: Kuka-IIWA correctly chooses the target block indicated by the blue arrow in simulator. The green vertical line indicates the location of the gripper tip on the table top. This line is invisible by the robot and is only for visualization purposes. }
    \label{fig:sim_results}
    \vspace{-15pt}
\end{figure}

\subsection{Results}
The following subsections outline our quantitative and qualitative assessment of the system performance. We examine the learning ability, task performance, and sensitivity of our proposed system. For quantitative measures, we conduct a thorough ablation study on the auxiliary losses used, the attention mechanism designed to guide the sound interpreter.

\subsubsection{Embodied learning}

We use two experiments to show that (1) our robot is able to learn abstract concept and this concept can evolve, and (2) our embodied agent understands the meaning of the sounds within the context. 

From the experiments on the Mix dataset, we show that our model can map two sounds to one object. From Fig. \ref{fig:turtlebot_result} middle right and Table \ref{tab:dataset_testing}, the training curve and success rate of our model on Mix dataset are similar compared with other single-class datasets in both environments. This result suggests that the model successfully learns to associate different sounds, such as ``dog'' and ``dog barking'' to the same concept. 

From the experiments on Wordset2 $\rightarrow$ Mix, we show that our model's interpretations of sound evolve dynamically. Fig. \ref{fig:turtlebot_result} right shows the training curve of our model for the Wordset2 $\rightarrow$ Mix experiment. As shown in Fig. \ref{fig:turtlebot_result} right and Table \ref{tab:dataset_testing}, after adding two new words on the top of our model's existing sound interpretations, although the reward drops slightly at first, it is able to quickly adapt to the new sound to meaning mappings by achieving similar reward and success rate as before. The success rates for this experiment reported in table \ref{tab:dataset_testing} are achieved after an additional timesteps of $0.5\times 10^6$ for the TurtleBot environment and $2\times 10^6$ for the Kuka environment. The additional number of timesteps needed to learn the new sounds for both environments are much less than training the networks from scratch.

\subsubsection{General to sound type and robotic tasks} We use all three types of sound data in our experiments: Wordset1, Wordset2, notes from NSynth dataset, environmental sounds from UrbanSound8K dataset, and Mix dataset. From the training curves (Fig. \ref{fig:turtlebot_result} (middle right)) and testing results (Table \ref{tab:dataset_testing}), we found that the performance of models with distinct sound datasets exhibit very little difference in training and testing. Therefore, our model (1) easily generalizes to different types of sound sources, whereas the general-purpose ASR systems can only translate speech sound signals, and (2) can capture the information of non-speech sound signals, which is lost when traditional voice-controlled robots only take translated text as inputs. 

Combining 1) and 2), our model is highly customizable to different applications as it adapts to different sound-to-meaning mappings, sound types, and robotic tasks with minimal changes. In contrast, the models of traditional voice-controlled robots would need to be redesigned for any of these changes to take place. 

\subsubsection{Auxiliary losses} We analyze the performance of models with auxiliary losses ablated, from both training and testing perspective.\footnote{The results of Wordset1, Wordset2, and NSynth are similar in ablations of auxiliary losses, so we take Wordset1 as an example to analyze.} 
We find that $\Lsound$ (gray in Fig. \ref{fig:turtlebot_result} left), helps feature extraction for the sound interpreter. 
For the Kuka task, the model with $\Lsound$ can achieve 8\% higher success rate at the test time (Table \ref{tab:auxiliary_testing}), meaning that the agent learns more robust features from $\Lsound$. For the more complex TurtleBot environment, the $\Lsound$  provides a guidance for sound interpretation, which eases the process of policy search. 

For the TurtleBot task, removing $\Ltarget+\Lobject$ (olive) or $\Lobject$ (orange) in Fig. \ref{fig:turtlebot_result} left, does not significantly affect training.
The loss's impact is partially handled by the pretrained weights, which also minimize $\Ltarget+\Lobject$.
However, the success rate is reduced for $\sim$8\% in testing (Table \ref{tab:auxiliary_testing}). 
We believe there are two reasons for this: (1) When the robot encounters new situations unseen by the pretraining data, it cannot draw guidance from $\Lobject$. (2) The concept vector from real sound data is distinctive from the one-hot vector used in pretraining. The agent cannot rebuild the association between the sound command and the corresponding object in the arena using $\Ltarget$. 
Removing $\Ltarget$ (brown) alone results in underfitting because the agent cannot choose the correct target. Thus, the robot is biased towards reaching the first object insight to minimize $\Lobject$. Therefore, optimizing all three auxiliary losses are essential for TurtleBot to achieve its goal.

\begin{table}[ht]
  \begin{center}
    \caption{Testing results of our model with ablations of auxiliary losses and attention layers trained with Wordset1}
    \label{tab:auxiliary_testing}
    \begin{tabular}{c l c} 
    \toprule
     \textbf{Environment} & \textbf{Network} & \textbf{Success rate} \\
      \midrule
      \multirow{8}{*}{TurtleBot} 
      & Attention$+\Lobject + \Ltarget + \Lsound$ & 0.920 \\
      & Attention $+ \Lobject + \Ltarget$ & 0.135 \\ 
      & Attention $+ \Ltarget + \Lsound$ &  0.885\\ 
      & Attention $+ \Lobject + \Lsound$ &  0.365\\ 
      & Attention $+ \Lsound$ &  0.845\\ 
      & Attention &  0.185\\ 
      & No attention $+\Lobject + \Ltarget + \Lsound$ & 0.805 \\ 
      & Random walk & 0.190\\
      \hline
       \multirow{4}{*}{Kuka} 
       & Attention $+\Lsound$ &  0.945\\
       & Attention & 0.870 \\ 
       & No attention$+\Lsound$ &  0.850\\ 
       & Random walk & 0.000\\
       \bottomrule
    \end{tabular}
  \end{center}
  \vspace{-5pt}
\end{table}

\begin{table}[ht]
  \begin{center}
    \caption{Testing results of our model in both environments with different types of sounds}
    \label{tab:dataset_testing}
    \begin{tabular}{c l c} 
    \toprule
    \textbf{Environment} & \multicolumn{1}{c}{\textbf{Dataset}} & \textbf{Success rate} \\
      \midrule

      \multirow{6}{*}{TurtleBot} 
      & Wordset1 & 0.920 \\
      & Wordset2 & 0.895 \\
      & NSynth & 0.950\\
      & UrbanSound8K & 0.930 \\
      & Mix & 0.870 \\ 
      & Wordset2 $\rightarrow$ Mix & 0.895 \\ 
     
      \hline
       \multirow{6}{*}{Kuka} 
       & Wordset1 & 0.945 \\
       & Wordset2 & 0.915 \\
       & NSynth & 0.910\\
       & UrbanSound8K & 0.875 \\
       & Mix & 0.945 \\ 
       & Wordset2 $\rightarrow$ Mix & 0.930 \\ 
       \bottomrule
    \end{tabular}
  \end{center}
  \vspace{-15pt}
\end{table}

\begin{figure*}[ht]
    \centering
    \includegraphics[scale = 0.4]{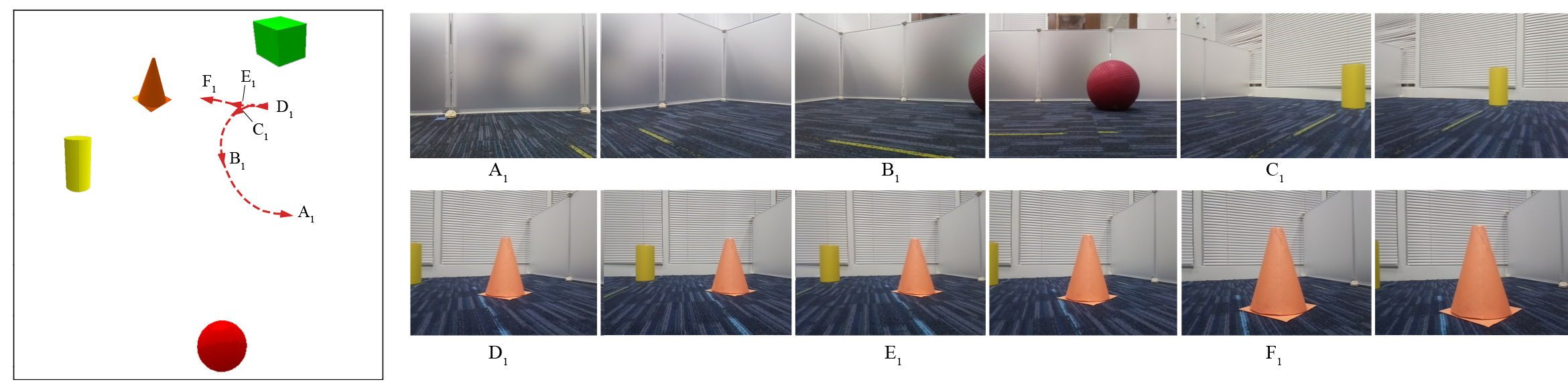}
    \caption{\textbf{Policy execution of our model in a real-world experiment}. The target is the cone. \textit{Left}: Navigation map of TurtleBot controlled by our model. The arrows indicate the orientation of the TurtleBot. \textit{Right}: The same episode in the TurtleBot's camera view. $A_1$-$F_1$ are the checkpoints in the trajectory corresponding to the labels in the left subfigure.}
    \label{fig:example_episode}
    \vspace{-10pt}
\end{figure*}

\subsubsection{Attention vs. no attention} To examine the effectiveness of the attention layer over the BiLSTM, we remove the attention layer from the network and train the network with all three types of sound for both environments. The training curves in Fig. \ref{fig:turtlebot_result} (middle left) shows that the removal of the attention layer negatively affects the training. Table \ref{tab:auxiliary_testing} shows that models with attention exhibit 8\%-12\% higher performance than those without attention,  which validates the necessity of attention layers. However, for NSynth dataset, attention layers do not make a difference because pure tone signals have simpler and cleaner patterns, which is easier for our network to learn.

\subsubsection{Qualitative simulation results} Fig. \ref{fig:sim_results} shows the snapshots of the two robots successfully reaching their goals. More demonstrations can be found in the video attachment.

\section{Real-world experiments}
\label{sec:real_exp}
We evaluate our trained model's performance on a real TurtleBot3 Waffle Pi. We assign ``house'' to the cube, ``dog'' to the sphere, ``bird'' to the cone, and ``tree'' to the cylinder. 
To narrow the gap between rendered and real-world images, we use domain randomization to learn robust visual features~\cite{Tobin_2017}. We randomize the color and texture of the four objects, the arena wall, and the background using the textures from the Describable Textures Dataset during training~\cite{cimpoi14describing}. 
Our model trained with domain randomization can be transferred to the real TurtleBot3 smoothly without further training on real world data.

In the real-world experiment, the communication between the TurtleBot3 and its host computer is established by the Robot Operating System (ROS). At every timestep, the trained agent receives an image from robot's RGB camera and the robot state vector calculated from odometry data and then produces an action. An episode lasts 12s. The wave files from the sound datasets are used as the sound input. We run a total of 40 tests (10 for each word) and count the number of times the robot approaches the correct target object.

The results of show that the success rates for each word are all 90.0\%, which indicates that the simulation to real world transfer is successful, even though the size, shape, and texture of the objects in the real world are slightly different from these in the simulation. The agent fails usually when the target object is not recognized. In such cases, the agent still actively searches for the target object but fails at approaching. 

A typical example episode is shown in Fig. \ref{fig:example_episode}. Our agent tends to find the target object with natural searching behavior. The robot usually first steps backward and rotates to look around for the target object. Once identified, it approaches towards the target until it is close to the object.

\section{Conclusion and Future work}
\label{sec:conclusion}

Inspired by human sound interpretation, we propose a cognitive end-to-end network for sound interpretations in vision-based robot control tasks. Our network integrates the image features from CNN, the sound features from a BiLSTM with attention, and the states of the robot to learn a policy using deep RL. We demonstrate our approach on a TurtleBot and a Kuka arm, and transferred the trained model to a real TurtleBot. Our experiments show promising results in both simulator and real world. Possible directions to explore in future work include: (1) generalizing the sound commands from single word to phrases and sentences, (2) using sound as feedback to our model, and (3) using sound interpretation as a building block towards robot language acquisition and human-robot interaction.

\section{Acknowledgements}
We thank Saurabh Gupta for thoughtful discussions and feedback on paper drafts.




\newpage
\clearpage
\bibliographystyle{IEEEtran}
\bibliography{BibFile}

\begin{thebibliography}{10}
\providecommand{\url}[1]{#1}
\csname url@rmstyle\endcsname
\providecommand{\newblock}{\relax}
\providecommand{\bibinfo}[2]{#2}
\providecommand\BIBentrySTDinterwordspacing{\spaceskip=0pt\relax}
\providecommand\BIBentryALTinterwordstretchfactor{4}
\providecommand\BIBentryALTinterwordspacing{\spaceskip=\fontdimen2\font plus
\BIBentryALTinterwordstretchfactor\fontdimen3\font minus
  \fontdimen4\font\relax}
\providecommand\BIBforeignlanguage[2]{{%
\expandafter\ifx\csname l@#1\endcsname\relax
\typeout{** WARNING: IEEEtran.bst: No hyphenation pattern has been}%
\typeout{** loaded for the language `#1'. Using the pattern for}%
\typeout{** the default language instead.}%
\else
\language=\csname l@#1\endcsname
\fi
#2}}

\bibitem{kocagoncu2017decoding}
E.~Kocagoncu, A.~Clarke, B.~J. Devereux, and L.~K. Tyler, ``Decoding the
  cortical dynamics of sound-meaning mapping,'' \emph{Journal of Neuroscience},
  vol.~37, no.~5, pp. 1312--1319, 2017.

\bibitem{hickok2000towards}
G.~Hickok and D.~Poeppel, ``Towards a functional neuroanatomy of speech
  perception,'' \emph{Trends in cognitive sciences}, vol.~4, no.~4, pp.
  131--138, 2000.

\bibitem{zhuang2014optimally}
J.~Zhuang, L.~K. Tyler, B.~Randall, E.~A. Stamatakis, and W.~D. Marslen-Wilson,
  ``Optimally efficient neural systems for processing spoken language,''
  \emph{Cerebral Cortex}, vol.~24, no.~4, pp. 908--918, 2014.

\bibitem{vanzo2016robust}
A.~Vanzo, D.~Croce, E.~Bastianelli, R.~Basili, and D.~Nardi, ``Robust spoken
  language understanding for house service robots,'' \emph{Polibits}, no.~54,
  pp. 11--16, 2016.

\bibitem{hermann2017grounded}
K.~M. Hermann, F.~Hill, S.~Green, F.~Wang, R.~Faulkner, H.~Soyer,
  D.~Szepesvari, W.~M. Czarnecki, M.~Jaderberg, D.~Teplyashin, \emph{et~al.},
  ``Grounded language learning in a simulated 3d world,'' \emph{arXiv preprint
  arXiv:1706.06551}, 2017.

\bibitem{smith2005development}
L.~Smith and M.~Gasser, ``The development of embodied cognition: Six lessons
  from babies,'' \emph{Artificial life}, vol.~11, no. 1-2, pp. 13--29, 2005.

\bibitem{wellsby2014developing}
M.~Wellsby and P.~M. Pexman, ``Developing embodied cognition: Insights from
  children’s concepts and language processing,'' \emph{Frontiers in
  psychology}, vol.~5, p. 506, 2014.

\bibitem{meltzoff1990towards}
A.~N. Meltzoff, ``Towards a developmental cognitive science: The implications
  of cross-modal matching and imitation for the development of representation
  and memory in infancy a,'' \emph{Annals of the New York academy of sciences},
  vol. 608, no.~1, pp. 1--37, 1990.

\bibitem{cruz2016multi}
F.~Cruz, G.~I. Parisi, J.~Twiefel, and S.~Wermter, ``Multi-modal integration of
  dynamic audiovisual patterns for an interactive reinforcement learning
  scenario,'' in \emph{IEEE/RSJ International Conference on Intelligent Robots
  and Systems (IROS)}, 2016, pp. 759--766.

\bibitem{fernandez2016natural}
R.~A.~S. Fernandez, J.~L. Sanchez-Lopez, C.~Sampedro, H.~Bavle, M.~Molina, and
  P.~Campoy, ``Natural user interfaces for human-drone multi-modal
  interaction,'' in \emph{IEEE International Conference on Unmanned Aircraft
  Systems (ICUAS)}, 2016, pp. 1013--1022.

\bibitem{burger2012two}
B.~Burger, I.~Ferran{\'e}, F.~Lerasle, and G.~Infantes, ``Two-handed gesture
  recognition and fusion with speech to command a robot,'' \emph{Autonomous
  Robots}, vol.~32, no.~2, pp. 129--147, 2012.

\bibitem{bastianelli2016discriminative}
E.~Bastianelli, D.~Croce, A.~Vanzo, R.~Basili, and D.~Nardi, ``A discriminative
  approach to grounded spoken language understanding in interactive robotics.''
  in \emph{International Joint Conference on Artificial Intelligence (IJCAI)},
  2016, pp. 2747--2753.

\bibitem{stramandinoli2016grounding}
F.~Stramandinoli, V.~Tikhanoff, U.~Pattacini, and F.~Nori, ``Grounding speech
  utterances in robotics affordances: An embodied statistical language model,''
  in \emph{Joint IEEE International Conference on Development and Learning and
  Epigenetic Robotics (ICDL-EpiRob)}, 2016, pp. 79--86.

\bibitem{paul2018efficient}
R.~Paul, J.~Arkin, D.~Aksaray, N.~Roy, and T.~M. Howard, ``Efficient grounding
  of abstract spatial concepts for natural language interaction with robot
  platforms,'' \emph{The International Journal of Robotics Research}, vol.~37,
  no.~10, pp. 1269--1299, 2018.

\bibitem{ovchinnikova2015multi}
E.~Ovchinnikova, M.~Wachter, V.~Wittenbeck, and T.~Asfour, ``Multi-purpose
  natural language understanding linked to sensorimotor experience in humanoid
  robots,'' in \emph{IEEE-RAS 15th International Conference on Humanoid Robots
  (Humanoids)}, 2015, pp. 365--372.

\bibitem{liu2019review}
R.~Liu and X.~Zhang, ``A review of methodologies for
  natural-language-facilitated human--robot cooperation,'' \emph{International
  Journal of Advanced Robotic Systems}, vol.~16, no.~3, 2019.

\bibitem{twiefel2014improving}
J.~Twiefel, T.~Baumann, S.~Heinrich, and S.~Wermter, ``Improving
  domain-independent cloud-based speech recognition with domain-dependent
  phonetic post-processing,'' in \emph{Twenty-Eighth AAAI Conference on
  Artificial Intelligence (AAAI)}, 2014.

\bibitem{anderson2018vision}
P.~Anderson, Q.~Wu, D.~Teney, J.~Bruce, M.~Johnson, N.~S{\"u}nderhauf, I.~Reid,
  S.~Gould, and A.~van~den Hengel, ``Vision-and-language navigation:
  Interpreting visually-grounded navigation instructions in real
  environments,'' in \emph{IEEE Computer Society Conference on Computer Vision
  and Pattern Recognition (CVPR)}, 2018, pp. 3674--3683.

\bibitem{chen2019touchdown}
H.~Chen, A.~Suhr, D.~Misra, N.~Snavely, and Y.~Artzi, ``Touchdown: Natural
  language navigation and spatial reasoning in visual street environments,'' in
  \emph{IEEE Computer Society Conference on Computer Vision and Pattern
  Recognition (CVPR)}, 2019, pp. 12\,538--12\,547.

\bibitem{yu2018interactive}
H.~Yu, H.~Zhang, and W.~Xu, ``Interactive grounded language acquisition and
  generalization in a 2d world,'' in \emph{International Conference on Learning
  Representations (ICLR)}, 2018.

\bibitem{chaplot2018gated}
D.~S. Chaplot, K.~M. Sathyendra, R.~K. Pasumarthi, D.~Rajagopal, and
  R.~Salakhutdinov, ``Gated-attention architectures for task-oriented language
  grounding,'' in \emph{Thirty-Second AAAI Conference on Artificial
  Intelligence (AAAI)}, 2018, pp. 2819--2826.

\bibitem{liang2019making}
H.~Liang, S.~Li, X.~Ma, N.~Hendrich, T.~Gerkmann, and J.~Zhang, ``Making sense
  of audio vibration for liquid height estimation in robotic pouring,''
  \emph{arXiv preprint arXiv:1903.00650}, 2019.

\bibitem{lathuiliere2019neural}
S.~Lathuili{\`e}re, B.~Mass{\'e}, P.~Mesejo, and R.~Horaud, ``Neural network
  based reinforcement learning for audio--visual gaze control in human--robot
  interaction,'' \emph{Pattern Recognition Letters}, vol. 118, pp. 61--71,
  2019.

\bibitem{zhang2001grounded}
Y.~Zhang and J.~Weng, ``Grounded auditory development by a developmental
  robot,'' in \emph{International Joint Conference on Neural Networks (IJCNN)},
  vol.~2, 2001, pp. 1059--1064.

\bibitem{liu2001robot}
Q.~Liu, S.~Levinson, Y.~Wu, and T.~Huang, ``Robot speech learning via entropy
  guided lvq and memory association,'' in \emph{International Joint Conference
  on Neural Networks (IJCNN)}, vol.~3, 2001, pp. 2176--2181.

\bibitem{manoonpong2005neural}
P.~Manoonpong, F.~Pasemann, J.~Fischer, and H.~Roth, ``Neural processing of
  auditory signals and modular neural control for sound tropism of walking
  machines,'' \emph{International Journal of Advanced Robotic Systems}, vol.~2,
  no.~3, p.~22, 2005.

\bibitem{davis1980comparison}
S.~Davis and P.~Mermelstein, ``Comparison of parametric representations for
  monosyllabic word recognition in continuously spoken sentences,'' \emph{IEEE
  transactions on acoustics, speech, and signal processing}, vol.~28, no.~4,
  pp. 357--366, 1980.

\bibitem{schuster1997bidirectional}
M.~Schuster and K.~K. Paliwal, ``Bidirectional recurrent neural networks,''
  \emph{IEEE Transactions on Signal Processing}, vol.~45, no.~11, pp.
  2673--2681, 1997.

\bibitem{chan2015listen}
W.~Chan, N.~Jaitly, Q.~V. Le, and O.~Vinyals, ``Listen, attend and spell,'' in
  \emph{2016 IEEE International Conference on Acoustics, Speech and Signal
  Processing (ICASSP)}, 2016, pp. 4960--4964.

\bibitem{bahdanau2014neural}
D.~Bahdanau, K.~Cho, and Y.~Bengio, ``Neural machine translation by jointly
  learning to align and translate,'' in \emph{International Conference on
  Learning Representations (ICLR)}, 2015.

\bibitem{luong2015effective}
M.-T. Luong, H.~Pham, and C.~D. Manning, ``Effective approaches to
  attention-based neural machine translation,'' in \emph{Conference on
  Empirical Methods in Natural Language Processing (EMNLP)}, 2015, pp.
  1412--1421.

\bibitem{vaswani2017attention}
A.~Vaswani, N.~Shazeer, N.~Parmar, J.~Uszkoreit, L.~Jones, A.~N. Gomez,
  {\L}.~Kaiser, and I.~Polosukhin, ``Attention is all you need,'' in
  \emph{Advances in Neural Information Processing Systems (NIPS)}, 2017, pp.
  5998--6008.

\bibitem{simonyan2014very}
K.~Simonyan and A.~Zisserman, ``Very deep convolutional networks for
  large-scale image recognition,'' in \emph{International Conference on
  Learning Representations (ICLR)}, 2015.

\bibitem{schulman2017proximal}
J.~Schulman, F.~Wolski, P.~Dhariwal, A.~Radford, and O.~Klimov, ``Proximal
  policy optimization algorithms,'' \emph{arXiv preprint arXiv:1707.06347},
  2017.

\bibitem{baselines}
P.~Dhariwal, C.~Hesse, O.~Klimov, A.~Nichol, M.~Plappert, A.~Radford,
  J.~Schulman, S.~Sidor, Y.~Wu, and P.~Zhokhov, ``Openai baselines,''
  \url{https://github.com/openai/baselines}, 2017.

\bibitem{warden2018speech}
P.~Warden, ``Speech commands: A dataset for limited-vocabulary speech
  recognition,'' \emph{arXiv preprint arXiv:1804.03209}, 2018.

\bibitem{Salamon:UrbanSound:ACMMM:14}
J.~Salamon, C.~Jacoby, and J.~P. Bello, ``A dataset and taxonomy for urban
  sound research,'' in \emph{International Conference on Multimedia (ACM-MM)},
  Orlando, FL, USA, Nov. 2014, pp. 1041--1044.

\bibitem{nsynth2017}
J.~Engel, C.~Resnick, A.~Roberts, S.~Dieleman, D.~Eck, K.~Simonyan, and
  M.~Norouzi, ``Neural audio synthesis of musical notes with wavenet
  autoencoders,'' in \emph{International Conference on Machine Learning
  (ICML)}, 2017, pp. 1068--1077.

\bibitem{pythonSpeechFeatures}
J.~lyons, ``Python speech features,''
  \url{https://github.com/jameslyons/python_speech_features}, 2013--2019.

\bibitem{coumans2019}
E.~Coumans and Y.~Bai, ``Pybullet, a python module for physics simulation for
  games, robotics and machine learning,'' \url{http://pybullet.org},
  2016--2019.

\bibitem{Tobin_2017}
J.~Tobin, R.~Fong, A.~Ray, J.~Schneider, W.~Zaremba, and P.~Abbeel, ``Domain
  randomization for transferring deep neural networks from simulation to the
  real world,'' \emph{IEEE/RSJ International Conference on Intelligent Robots
  and Systems (IROS)}, pp. 23--30, 2017.

\bibitem{cimpoi14describing}
M.~Cimpoi, S.~Maji, I.~Kokkinos, S.~Mohamed, and A.~Vedaldi, ``Describing
  textures in the wild,'' in \emph{IEEE Computer Society Conference on Computer
  Vision and Pattern Recognition (CVPR)}, 2014.

\end{thebibliography}
\clearpage
\end{document}